\documentclass[11pt]{article}

\usepackage[T1]{fontenc}
\usepackage[utf8]{inputenc}

\usepackage[a4paper,margin=1in]{geometry}

\usepackage{amsmath,amssymb}  % load math first
\usepackage{mathptmx}         % Times text + math
\usepackage[scaled=0.9]{helvet}  % Helvetica for \textsf (headers)
\usepackage{courier}          % Courier for \texttt (code spans)
\usepackage{textcomp}         % extra text symbols

\usepackage{microtype}
\usepackage[english]{babel}
\usepackage{csquotes}
\usepackage{enumitem}
\setlist{topsep=4pt,itemsep=2pt,parsep=0pt}

\usepackage{xcolor}
\usepackage[colorlinks=true,linkcolor=blue!55!black,citecolor=blue!55!black,urlcolor=blue!55!black,filecolor=blue!55!black]{hyperref}
\usepackage{url}
\usepackage{bookmark}

\usepackage{graphicx}
\usepackage{longtable}
\usepackage{booktabs}
\usepackage{array}
\usepackage{fancyhdr}
\usepackage{titlesec}
\titleformat{\section}{\normalfont\large\bfseries}{}{0pt}{}
\titleformat{\subsection}{\normalfont\normalsize\bfseries}{}{0pt}{}
\titlespacing*{\section}{0pt}{1.4em}{0.6em}
\titlespacing*{\subsection}{0pt}{1.1em}{0.4em}

\title{\Large\textbf{How to Build Marcus's Algebraic Mind: From Thagard's \emph{Brain--Mind} Viewpoint}\\[0.6em]
\large\normalfont An Algebro-Deterministic Substrate over Galois Fields as the\\
Reversible, Auditable Binding that Statistical Minds Structurally Lack}

\author{Hiroyuki Chuma$^{1}$ \quad Kanji Otsuka$^{2}$ \quad Yoichi Sato$^{3}$\\[0.4em]
\normalsize\itshape
$^{1}$Institute of Innovation Research, Hitotsubashi University (Professor Emeritus) \\
$^{2}$Meisei University (Professor Emeritus) \quad
$^{3}$Shuhari System}

\date{\normalsize\itshape Draft revision of arXiv:2605.21379. Target venue: cs.NE / cs.AI.\\[0.3em]
\normalsize\upshape \today\ (revised edition)}

\begin{document}

\maketitle
\thispagestyle{empty}

\vspace{-1.5em}
\begin{center}\rule{0.5\linewidth}{0.5pt}\end{center}
\vspace{0.5em}

\hypertarget{abstract}{%
\section*{Abstract}\label{abstract}}

This paper reports a convergence that neither of the two programs it
joins was looking for. In \emph{The Algebraic Mind}, Marcus {[}17{]}
identified three components any adequate architecture must support ---
operations over variables, recursively structured representations, and a
representation of individuals distinct from kinds --- and showed that
the standard multilayer perceptron supports none of them, closing with
the explicit admission that a \emph{neural} implementation, organized
around registers and treelets and built by developmental cascades rather
than gradient descent, remained a programmatic conjecture. In
\emph{Brain--Mind}, Thagard {[}31{]}, building on Eliasmith's Semantic
Pointer Architecture {[}30{]}, ran the argument from the other end: he
took \emph{binding} as the single mechanism from which the full
architecture of mind is assembled --- perception, language, emotion,
consciousness, and the self --- and thereby made a specific binding
algebra (circular convolution) load-bearing for the whole of higher
cognition. Marcus specifies \emph{what} an adequate substrate must do
and leaves the algebra of its registers open; Thagard supplies a binding
algebra and extends it to consciousness and self, but that algebra is
lossy and degrades under exactly the deep recursion his own account of
the self and the emotions demands. Neither author was writing about
hardware, and the two never cite each other; yet the operation each
needs is the same operation, and it is the one a memory architecture
built for speed, power, and cost on commodity silicon turns out to have.

VaCoAl {[}5{]} is a hyperdimensional computing \emph{architecture}
organized end-to-end around one algebraic primitive --- XOR-and-shift
over GF(2), realized by primitive-polynomial linear-feedback shift
registers --- and PyVaCoAl is its substantially extended software
realization, the system in which the empirical results of this paper are
obtained: it materializes the multi-stage rescue circuit, scales toward
million-dimensional operation, and introduces the Rescue-Rate phase
transition that gives the architecture its effective identity. VaCoAl's
intended physical substrate is not merely a design target: it already
exists as CASRAM, an SRAM/DRAM content-addressable-memory realization
commercialized by Shuhari System {[}34{]}. The two are related as design
to substantially extended realization, not as design to product; nothing
in this paper's empirical record derives from CASRAM, and direct
hardware benchmarking --- the ultra-low-power, nanosecond-scale claim
specifically --- remains a pending measurement, now the object of an
author--Shuhari System collaboration, rather than an unbuilt hypothesis.
Binding, \texttt{Bind(R,\ F)\ =\ R\ \ensuremath{\oplus}\ shift(F)}, is the shared
algebraic core: \emph{exactly} reversible, non-commutative under shift,
and defined uniformly over all fillers; it fills Marcus's open algebra
of registers (Pillar 1), supports non-commutative compositional bundling
at fixed dimension (Pillar 2), and separates individuals from kinds
under the same algebra (Pillar 3) --- while, read through Thagard, it
removes the depth-degradation that afflicts convolution precisely where
his recursive bindings (an emotion binds a self; the self is itself
bound; an intention binds both) are deepest.

We make three scoped claims and are careful about each.
\emph{Capability}: exact reversible binding at O(N) yields compositional
generalization together with post-hoc auditability, a combination no
lossy or learned substrate offers. \emph{Necessity}: that two
independent cognitive-architecture programs --- Marcus's
symbol-manipulation pillars and Thagard's binding-based mind --- and a
biological circuit (dentate gyrus--CA3 {[}6{]}) all require the
\emph{same} reversible-compositional algebra is evidence, via convergent
evolution rather than biomimicry, that the algebra is a
substrate-independent requirement, not a design preference. A second and
sharper form of the necessity claim is reported here for the first time:
the architecture's own discriminating power and its tolerance of failure
are the same phenomenon. Repair every block collision --- set
\texttt{RR\ =\ 1} --- and the system becomes bit-identical to a hash
dictionary, at which point every reachable candidate is indistinguishable
from every other and the confidence path-integral collapses. A memory
that never fails keeps no record of which routes were difficult, and
therefore has no basis on which to prefer one. \emph{Position}: we do not
claim to surpass large language models. The substrate is
\emph{orthogonal} to them --- it supplies the reversible, auditable,
multi-hop relational reasoning that statistical embeddings structurally
lack, and makes no claim on the fluency at which those embeddings excel.
Thagard's account of consciousness and the self enters not as a theory we
implement but as an independent \emph{witness} that the layer we supply
is one intelligence was, on at least one substrate, compelled to build.

\emph{Not claimed.} We do not implement consciousness, do not adjudicate
Thagard's theory against rivals, and report no cognitive experiment:
Marcus's own case studies have not been re-run on the architecture.
Bit-exact reversibility holds in silicon and only approximately under
biological noise. Every speed and power figure attached to VaCoAl is
architectural and, in this paper, unbenchmarked; every
reasoning-capability figure is a PyVaCoAl result obtained in the
\texttt{RR\ =\ 0}, sufficient-depth regime specifically, and does not
transfer to the hardware realization without a measurement we have not
yet made.

\textbf{Keywords:} algebraic mind, semantic pointers, variable binding,
hyperdimensional computing, Galois fields, XOR, treelets, convergent
evolution, reversible reasoning, complementarity with LLMs.

\begin{center}\rule{0.5\linewidth}{0.5pt}\end{center}

\hypertarget{introduction}{%
\section*{1. Introduction}\label{introduction}}

\hypertarget{two-diagnoses-of-one-absence}{%
\subsection*{1.1 Two diagnoses of one
absence}\label{two-diagnoses-of-one-absence}}

Marcus {[}17{]} and Thagard {[}31{]} rarely appear on the same page, yet
their programs are the two ends of one diagnosis. Marcus works from the
mechanics upward: he asks what elementary capacities a mind must have,
isolates three --- free generalization of operations over variables,
recovery of constituents from structured wholes, and the individual/kind
distinction --- and shows that the connectionist pattern associator, as
standardly used, has none of them. His book closes on an honest
promissory note: even granting the three pillars, \emph{how they are
implemented in neural hardware remained to be discovered}, with
registers, treelets, and gene-cascade development offered as plausible
but unproven conjecture.

Thagard works from the architecture downward. Taking Eliasmith's
semantic pointers {[}30{]} as the universal representational currency,
he makes \emph{binding} --- the composition of sensory, motor,
emotional, and verbal representations into compressed, dereferenceable
wholes --- the single mechanism from which he builds perception,
language, and then, in the book's most ambitious chapters, emotion,
consciousness, and the self. In Thagard's architecture an emotion is a
binding of a situation, a physiological reaction, an appraisal, and a
\emph{self}; the self is itself a binding of many neural
representations; an intention binds situation, evaluation, action, and
self again. Binding, and the competition among bound pointers, carries
the entire load.

Read together, the two programs name the same absent capacity from
opposite sides. Marcus: the connectionist mind cannot bind constituents
into structures and get them back. Thagard: binding is what a mind is
\emph{made of}, up to and including its consciousness of itself. What
neither program had, in 2001 or in 2019, was a binding operation that is
at once neurally plausible, computationally cheap, and --- the property
that turns out to matter most --- \emph{exactly reversible}. Marcus left
the algebra of registers open. Thagard adopted an algebra (circular
convolution) that is only approximately reversible and that degrades
with the depth of nesting his own theory of the self and the emotions
requires.

\hypertarget{what-is-and-is-not-claimed}{%
\subsection*{1.2 What is and is not
claimed}\label{what-is-and-is-not-claimed}}

Because this paper joins an engineering substrate to two cognitive
programs written without reference to hardware, it is worth stating at
the outset which claims are load-bearing and which are not. We ask the
reader to hold us to a narrow scope, because the argument is easy to
over-read.

\textbf{Capability.} Exact reversible binding over GF(2), computed at
O(N) cost and inverted exactly, supplies each of Marcus's three pillars
as an architectural primitive rather than as an emergent property of
training, and supplies with them a post-hoc auditability that no lossy
or learned substrate offers: any composite can be decomposed and its
constituents inspected in their roles. We claim that this combination
--- compose freely, then recover \emph{exactly}, at depth --- is
structurally unavailable to substrates whose inverse is approximate,
which can report a nearest neighbour but not a constituent.

\textbf{Necessity.} We claim the reversible-compositional layer is a
requirement rather than a preference, and we defend the claim twice, on
independent grounds. The first defence is convergence (\S{}8): two
cognitive-architecture programs derived from disjoint starting points,
plus one biological circuit, all arrive at the same algebra, under no
common design lineage at the level of solution. The second is internal
to the architecture (\S{}7): its ability to rank what it retrieves and its
tolerance of failure are not two features but one, so that repairing
every collision does not improve the system but silences it. Both
defences run in the same uncomfortable direction --- they say the layer
cannot be had by adding a component to something that lacks it.

\textbf{Position.} We do \textbf{not} claim that PyVaCoAl/VaCoAl is the
mind, that the brain is literally a shift register, or that the
architecture surpasses large language models. The layer this substrate
supplies is \emph{orthogonal} to the LLM --- it is what statistical
embeddings structurally lack --- so the contribution is a division of
labour, not a contest.

Orthogonality, not superiority, is the correct geometry, and it is not a
rhetorical choice: it is the geometry the architecture is built on.
Quasi-orthogonality in a high-dimensional binary space is the property
that lets the engine superpose without interference; we simply extend
the same relation to the engine's place beside the LLM. Two orthogonal
axes --- linguistic fluidity on one, reversible multi-hop logical
integrity on the other --- compose into a reasoning stack neither
completes alone.

\textbf{Not claimed.} We do \textbf{not} attempt to adjudicate Thagard's
theory of consciousness as cognitive science --- to show, by a coherence
comparison, that our mechanism explains the phenomena better than his.
That adjudication belongs to a different paper addressed to a different
community. We do not implement consciousness, and the hard problem is
untouched by the choice of binding operation. No cognitive experiment is
reported here: Marcus's Chapter 3 and Chapter 5 case studies have not
been re-run on the architecture, and doing so is listed in \S{}14 as a
test, not a result. Bit-exact reversibility holds in silicon and only
approximately under biological noise. And every claim about speed and
power belongs to the VaCoAl/CASRAM level and is unbenchmarked within
this paper, while every demonstrated reasoning result belongs to
PyVaCoAl and to one specific parameter regime; \S{}3.0 keeps the two apart
and \S{}14 restates the boundary.

\hypertarget{why-introspection-is-a-decomposition-problem}{%
\subsection*{1.3 Why introspection is a decomposition
problem}\label{why-introspection-is-a-decomposition-problem}}

Before the architecture, one illustration, because the property this
paper turns on is easy to mistake for an engineering nicety.

Consider what is being asked of a mind when it asks itself \emph{why did
I feel that?} On Thagard's own account the answer is not a lookup. An
emotion is a binding of a situation, a physiological reaction, an
appraisal, and a self; to answer the question is to take that binding
apart and inspect what occupies each role --- to find that the appraisal
slot held a judgment of threat, that the self slot held a version of
oneself that no longer obtains, that the situation slot held something
the appraisal had misread. The object retrieved is not the feeling; it is
the feeling's constituents in their roles. A mind that can only recover
the whole, or a point near the whole, can report that it felt something
and cannot report what the feeling was made of. It can be moved and
cannot be informed.

The same shape recurs wherever Thagard's architecture reaches furthest.
Revising a self-representation is a re-binding: it requires locating the
constituent to be changed, removing it, and putting another in its place
without disturbing the rest. Planning is the projection of a self into a
situation that does not obtain, which requires holding a factual binding
and a contrary-to-fact one side by side. Analogy is the transfer of a
system of relations, which requires that the relations survive retrieval
(\S{}9.1). None of these is a matter of recovering \emph{approximately} the
right thing. Each requires that the parts be genuinely separable, and an
architecture in which they are separable only to within an accumulating
error does not do them slightly worse. It does something else.

Two structural properties are required, and the engine supplies both.
\emph{Reversibility}: because
\texttt{Bind(R,\ F)\ =\ R\ \ensuremath{\oplus}\ shift(F)} is exactly invertible, one
can unbind at a site, inspect, substitute, and rebind, leaving every
other binding untouched. \emph{Determinism}: because the diffusion is
algebraic rather than sampled, the same composite is bit-exactly
reproducible, so a decomposition can be re-run and compared rather than
merely recollected. Section 13 shows that these two properties are also
precisely what Pearl's second and third rungs require, which is not a
coincidence: introspection and counterfactual reasoning are the same
operation pointed in different directions.

\hypertarget{why-thagards-viewpoint-reframes-the-marcus-program}{%
\subsection*{1.4 Why Thagard's viewpoint reframes the Marcus
program}\label{why-thagards-viewpoint-reframes-the-marcus-program}}

Reading Marcus through \emph{Brain--Mind} changes the status of the
three pillars. In Marcus's own framing the pillars are the endpoint: an
architecture is adequate if it supports them. In Thagard's framing they
become the \emph{floor} --- the substrate-level prerequisites for a mind
that goes on to feel, to be conscious, and to represent itself.
Operations over variables, structured representations, and the
individual/kind distinction are exactly the capacities a binding-based
architecture must already have before it can bind a self into an emotion
or compete pointers into consciousness. The Thagard viewpoint therefore
does not compete with the Marcus reading; it tells us \emph{what the
Marcus substrate is for}, and it relocates the payoff of exact
reversibility to where Thagard's constructions are deepest and
convolution is weakest --- the recursive nesting of self, emotion, and
intention (\S{}5.2, \S{}9).

It also supplies the argument, which the Marcus paper on its own could
not make, that the reversible-compositional layer is a \emph{necessity}
rather than one design choice among many. That argument is convergence,
and it is the spine of \S{}8.

\hypertarget{structure-of-the-paper}{%
\subsection*{1.5 Structure of the
paper}\label{structure-of-the-paper}}

Section 2 sets out Marcus's three pillars, Thagard's binding
architecture, and the gap they share. Section 3 introduces the substrate
at the level of operational commitments, and --- because the distinction
governs every claim that follows --- separates the base architecture,
its extended software realization, and its physical realization as
CASRAM. Sections 4--6 develop the correspondence pillar by pillar.

Section 7 is the core of the paper: it reports the Rescue-Rate phase
transition and isolates its general form, that the information needed to
prefer one retrieval over another is generated only as a by-product of
failing. Section 8 develops the convergence argument, and \S{}8.2 states
the discipline --- we borrowed the problem, not the solution --- that
separates a convergence claim from biomimicry. Section 9 reads Thagard's
higher cognition on the substrate as witness rather than implementation,
with analogy treated in \S{}9.1. Section 10 condenses the biological
reading; Section 11 compares the result with the 2001 alternatives and
the 2024 default. Section 12 draws the consequence for output format:
that the natural product of this architecture is a family of structures
indexed by resolution rather than a single answer. Section 13 extends
the substrate to Pearl's ladder and restates the CASRAM boundary;
Section 14 gives reservations and testable predictions, including one
that would refute the necessity claim. Section 15 concludes by carrying
two consequences out of the paper separately from the architecture that
produced them. Appendix A states precisely what kind of equivalence is
being claimed between silicon and tissue, and what kind is not.

\begin{center}\rule{0.5\linewidth}{0.5pt}\end{center}

\hypertarget{marcuss-three-pillars-thagards-binding-architecture-and-the-shared-gap}{%
\section*{2. Marcus's three pillars, Thagard's binding architecture, and
the shared
gap}\label{marcuss-three-pillars-thagards-binding-architecture-and-the-shared-gap}}

\hypertarget{marcuss-three-pillars-and-open-questions}{%
\subsection*{2.1 Marcus's three pillars and open
questions}\label{marcuss-three-pillars-and-open-questions}}

Marcus {[}17{]} reconstructs symbol manipulation as three separable
hypotheses. \textbf{Pillar 1 --- operations over variables:} the mind
represents abstract relationships between variables and generalizes them
freely to novel instances, as the inflection that adds \emph{-ed}
applies to any stem, including a pseudo-verb never encountered. The
technical content is that there must exist an operation defined
uniformly over all instantiations of a variable, not over a stored
sample --- a universally quantified mapping. \textbf{Pillar 2 ---
structured representations:} the mind represents complex structures so
that the structural relations among constituents are preserved and
recoverable; \emph{the book on the table} and \emph{the table on the
book} share constituents and roles yet must remain distinguishable and
decodable. \textbf{Pillar 3 --- individuals versus kinds:} the mind
distinguishes a representation of a specific individual from a
representation of the kind it instantiates, as object permanence and the
tracking of individuals through time require.

Marcus is explicit about what his program leaves open. \textbf{(A) The
algebra of operations on registers:} the treelet supplies a structural
scaffold but not the elementary operations under which Bind, Unbind, and
recursion are defined. \textbf{(B) The source of structural specificity
without a blueprint:} how finely specified microcircuitry arises when
the genome does not encode a point-by-point wiring diagram. \textbf{(C)
What a register looks like in tissue.} The engineering substrate below
answers (A); the companion biological reading (\S{}10) answers (B) and (C).
A silicon-level analogue of (C) --- what a register looks like in
hardware, as opposed to tissue --- already has an engineering answer:
the architecture is physically realized as CASRAM (\S{}3.0, \S{}13). We keep
this distinct from the biological reading; a commercial chip built to
the same algebra is an existence proof that the architecture is
buildable, not a second instance of Marcus's question about neural
tissue.

\hypertarget{thagards-semantic-pointer-architecture-and-where-it-strains}{%
\subsection*{2.2 Thagard's semantic-pointer architecture, and where it
strains}\label{thagards-semantic-pointer-architecture-and-where-it-strains}}

Thagard {[}31{]} adopts Eliasmith's semantic pointers {[}30{]}:
compressed neural representations formed by binding, functioning as
pointers that must be \emph{dereferenced} to recover what they compress.
On this base he builds a strikingly unified account --- an emotion binds
situation, physiology, appraisal, and self; consciousness is a
competition among semantic pointers for a limited broadcast; the self is
a multilevel binding of representations about oneself. The unifying
commitment is that one mechanism, binding-plus-competition, recurs at
every level of mind.

The strain is in the substrate, not the vision. SPA binds by circular
convolution {[}22{]}, whose unbinding is a \emph{quasi}-inverse: it
returns the filler plus a noise term that accumulates with the number
and depth of bindings, and it requires a clean-up memory to be usable at
all. Thagard's boldest constructions are also his most deeply nested ---
the self is bound into the emotion, and the self is itself a bound
composite, and an intention binds the emotion and the self again --- so
the very places where his theory reaches furthest are the places where a
depth-degrading substrate is least able to support it. Thagard invokes
systematicity and productivity --- the capacity to compose and decompose
unboundedly nested structure --- as requirements on cognition, while
resting them on an operation that loses fidelity with each level of
nesting. This is not a refutation of Thagard; it is the seam along which
a reversible substrate is exactly what his own account needs.

\hypertarget{the-shared-gap}{%
\subsection*{2.3 The shared gap}\label{the-shared-gap}}

Both programs, then, point at reversible binding. Marcus needs an exact
operation over variables and an exact recovery of constituents (Pillars
1--2) and leaves its algebra unspecified. Thagard needs binding that
survives deep recursion and adopts one that does not. And the dominant
contemporary substrate --- the learned dense embedding --- is worse than
either: its composition is a nonlinear superposition with no algebraic
inverse at all, so constituents fuse irreversibly and cannot be read
back. The gap the rest of this paper fills is a single reversible
binding algebra that closes Marcus's open question, relieves Thagard's
strain, and supplies what the statistical embedding structurally cannot
have. Table 1 states the trade each candidate substrate makes; the point
of the table is not that the alternatives got the algebra wrong but that
each buys one property at the price of another, and that the engine
declines that particular trade.

\begin{center}
\small
\begin{tabular}{@{}p{3.1cm}p{5.4cm}p{5.4cm}@{}}
\toprule
\textbf{Axis} & \textbf{Prior substrates} & \textbf{VaCoAl} \\
\midrule
Binding operation & Outer product; circular convolution; spike timing; attention-weighted mixture & XOR-and-shift over GF(2) \\
\addlinespace
Unbinding & Contraction (exact); quasi-inverse (lossy); re-synchronization (timing-bound); no inverse defined & Exact algebraic inverse \\
\addlinespace
Role asymmetry & Tensor order; permutation; firing order; learned position & Shift permutation, structural \\
\addlinespace
Cost of recursion & Multiplicative growth in dimension; fixed dimension with accumulating crosstalk; rising temporal precision; compounding read-out error & Fixed dimension; bounded by collision saturation \\
\addlinespace
Inspectability & Exact but unaffordable; approximate; timing-dependent; not defined & Bit-exact, auditable post hoc \\
\addlinespace
Reported failure & Dimensional explosion; crosstalk accumulation; precision wall; silent degradation & Block collision, reported as CR1/CR2 \\
\bottomrule
\end{tabular}
\end{center}

\begin{center}\small\emph{Table 1. What each substrate buys and what it pays. The rightmost
column is not a claim of superiority on every axis; it states which trade the
engine declines to make.}\end{center}

\begin{center}\rule{0.5\linewidth}{0.5pt}\end{center}

\hypertarget{the-vacoal-architecture-and-its-pyvacoal-realization}{%
\section*{3. The VaCoAl architecture and its PyVaCoAl
realization}\label{the-vacoal-architecture-and-its-pyvacoal-realization}}

\hypertarget{two-levels-and-which-claims-belong-to-which}{%
\subsection*{3.0 Two levels, and which claims belong to
which}\label{two-levels-and-which-claims-belong-to-which}}

Throughout this paper we keep three levels distinct, because conflating
them would misstate what is demonstrated, what is physically realized
but unmeasured in this record, and what is projected. \textbf{VaCoAl}
{[}5{]} is the \emph{base architecture}: the algebro-deterministic
Galois-field diffusion realized by primitive-polynomial LFSRs,
endogenous collision suppression through memory depth 2\ensuremath{{}^{m}},
quasi-orthogonal diffusion, and reversible XOR-and-shift binding. Its
intended physical home is an SRAM/DRAM content-addressable memory
performing in-memory matching at nanosecond scale under a 1/N
block-activation rule --- the source of the ultra-low-power claim. This
is not merely an intended target: the architecture already has a
physical realization, \textbf{CASRAM}, an SRAM/DRAM content-addressable
memory commercialized by Shuhari System. We stress the distinction this
paper actually needs: CASRAM's existence establishes that the
architecture is physically buildable, not that its performance is
measured here --- no direct CASRAM benchmark is reported in this paper,
and the low-power, nanosecond-scale claim remains a \textbf{pending
measurement} rather than an unbuilt hypothesis (see \S{}13).

\textbf{PyVaCoAl} {[}5{]} is not merely a port of that architecture but
a substantially \emph{extended} software realization, and it is the
system in which every empirical result of this program is obtained. It
runs as a \emph{de facto} DRAM-CAM on a CPU (10--15 ns access), forgoing
SRAM's \textasciitilde1 ns parallelism; it \textbf{fully implements} the
multi-stage collision-avoidance (rescue) circuit --- using the host
language's standard hash for block-collision resolution --- rather than
leaving it as a hardware specification; it scales the hypervector toward
one-million dimensions; it materializes only the parity region as an
address, the information region being indexed by label rather than
carried as bits; and it introduces the tunable Rescue Rate
\texttt{RR\ \ensuremath{\in}\ {[}0,1{]}} together with the memory-depth phase
transition that gives the architecture its \emph{effective identity}. At
\texttt{RR\ =\ 1} the output becomes bit-identical to a plain hash
dictionary and the confidence path-integral loses discriminating power;
only at \texttt{RR\ =\ 0} with sufficient depth does CR2 acquire the
semantic-selection behaviour on which the higher-level results depend.
Section 7 treats this transition not as a configuration detail but as
the paper's central finding about what a memory must give up in order to
be able to choose. The CR2 path integral, the Don't Care semantic
pruning with its STDP-like closed form, and the multi-predicate
multi-hop reasoning are all PyVaCoAl-level phenomena, demonstrated in
software; PyVaCoAl's measured throughput is therefore a \emph{lower
bound} on what the CASRAM realization of the same architecture should
achieve, not a substitute for measuring it.

The mapping to Marcus's pillars in \S{}\S{}4--6 rests on the \emph{shared
algebraic core} --- reversible XOR-and-shift binding, common to all
three levels. Where a claim depends on the rescue circuit, the RR phase
transition, CR2 selection, or million-dimensional scale, it is a
\textbf{PyVaCoAl (demonstrated)} claim; where it depends on in-memory
speed or power, it is a \textbf{VaCoAl/CASRAM (physically realized,
benchmark pending)} claim. We label them accordingly rather than
absorbing all three into one compound, and we do not let the existence
of CASRAM stand in for a measurement we have not made.

\hypertarget{engine-commitments}{%
\subsection*{3.1 Engine commitments}\label{engine-commitments}}

The three commitments below belong to the base architecture (C1--C2) and
to its realized readout (C3, materialized and measured in PyVaCoAl).
Stated at the level the comparison requires: \textbf{C1 ---
XOR-and-shift is the unique elementary primitive.} Every operation ---
diffusion, binding, unbinding, bundling --- decomposes into
XOR-and-shift over GF(2); the primitive is the same at every scale.
\textbf{C2 --- diffusion is deterministic and reversible.} For a
primitive polynomial \texttt{G(x)\ \ensuremath{\in}\ GF(2){[}x{]}} of degree m, the
diffusion \texttt{\ensuremath{\Psi}(P)\ =\ x\ensuremath{{}^{m}}P(x)\ mod\ G(x)} is implemented as m
shift-and-conditional-XOR steps of a linear-feedback shift register; \ensuremath{\Psi}
is GF(2)-linear, bijective on the non-collision input space, and
satisfies quasi-orthogonal diffusion --- for two inputs whose XOR
difference is not in the collision kernel, output Hamming distance
concentrates near m/2 with variance \ensuremath{\approx} m/4, exponentially tight in m.
\textbf{C3 --- readout is block-wise majority voting under bounded
confidence.} The length-L hypervector is partitioned into N blocks of
length q = L/N, each with its own polynomial \texttt{Gb(x)} and seed;
readout yields a per-step confidence ratio CR1 and a path-integral
CR2(n) = \ensuremath{\prod}\ensuremath{{}_{i}} CR1(i), a graded reliability metric under partial collision.

The three are not independent design decisions taken in sequence. C1
fixes the algebra; C2 is what that algebra does when iterated m times
inside a block; C3 is what a bank of such blocks does when their outputs
are counted. There is one decision in this architecture, and the rest is
arithmetic.

\hypertarget{why-xor-and-shift-and-not-multiplication-convolution-or-addition}{%
\subsection*{3.2 Why XOR-and-shift, and not multiplication, convolution,
or
addition}\label{why-xor-and-shift-and-not-multiplication-convolution-or-addition}}

The choice is forced, not stylistic, and the whole comparison turns on
it. GF(2) admits exactly two non-trivial binary operations: XOR (its
addition) and AND (its multiplication). XOR is its own inverse and is
reversible; AND destroys information and cannot be inverted from
outputs. Reversibility is precisely what licenses bit-exact recovery of
constituents from a bundle --- the property Marcus most wanted from
Pillar 2, and the property that circular convolution, with its
accumulating quasi-inverse noise, cannot deliver. There is no third
candidate to weigh: in a binary field the choice is between the
operation that keeps information and the operation that discards it.
Shift --- multiplication by x modulo G(x) --- supplies what XOR alone
cannot, non-commutativity: \texttt{Bind(R,\ F)\ =\ R\ \ensuremath{\oplus}\ shift(F)} differs from
\texttt{Bind(F,\ R)} because shift acts asymmetrically, which is what
lets a bundle distinguish \emph{the dog bites the man} from \emph{the
man bites the dog}. The asymmetry is a permutation, structurally present
before anything is trained and costing one shift; it is not a learned
weighting, a positional embedding, or a timing difference. And the
avalanche property of \ensuremath{\Psi} --- a single-bit
input change concentrates the output near Hamming weight m/2 ---
eliminates, at the level of the elementary operation, the
bounded-failure mode in which similar inputs yield similar outputs that
random sparse projection can only suppress statistically.

\hypertarget{the-elementary-operation-is-the-whole-argument}{%
\subsection*{3.3 The elementary operation is the whole
argument}\label{the-elementary-operation-is-the-whole-argument}}

Everything in \S{}\S{}4--6 is bookkeeping on
\texttt{Bind(R,\ F)\ =\ R\ \ensuremath{\oplus}\ shift(F)}. We flag this because the
three pillars are conventionally treated as three problems, and a reader
expecting three solutions will be looking for machinery that is not
there. Pillar 1 asks for an operation defined independently of the value
a variable holds; the bind is such an operation because F appears only
as an argument. Pillar 2 asks that structures built from identical
constituents be distinguishable and decomposable; the bind distinguishes
them because shift is asymmetric and decomposes them because XOR is an
involution. Pillar 3 asks that individuals be separable from kinds
without a second representational format; the bind separates them
because distinct roles are distinct keys into one shared space. One
operation, three pillars, no additional apparatus.

The parsimony is the only evidence of its kind this paper can offer, and
we do not want to overstate it. We cannot show that GF(2) is the algebra
the mind uses. What we can observe is that a specification written in
2001 without reference to any hardware, listing three requirements its
author took to be logically independent, turns out to be satisfiable by
one operation chosen for reasons that had nothing to do with cognition
--- and that the operation is the only one its field admits. That is not
proof. It is the kind of coincidence that usually indicates a structural
fact rather than an engineering accident, and Appendix A states how much
weight it will bear.

\begin{center}\rule{0.5\linewidth}{0.5pt}\end{center}

\hypertarget{pillar-1-operations-over-variables-as-reversible-xor-binding}{%
\section*{4. Pillar 1 --- operations over variables as reversible XOR
binding}\label{pillar-1-operations-over-variables-as-reversible-xor-binding}}

Marcus's Pillar 1 requires an operation defined uniformly over every
instantiation of a variable, with free generalization as its hallmark:
an inflection learned on a finite set extends without retraining to
pseudo-verbs. Made precise, the pillar asks for three things --- the
operation must be (i) definable independently of the value the variable
holds, (ii) invariant in its structural effect across substitutions, and
(iii) inspectable, so that the variable--value relation is recoverable
from the output.

VaCoAl's bind satisfies all three. (i) follows from linearity ---
\texttt{Bind(R,\ F\ensuremath{{}_{1}}\ \ensuremath{\oplus}\ F\ensuremath{{}_{2}})\ =\ Bind(R,\ F\ensuremath{{}_{1}})\ \ensuremath{\oplus}\ Bind(R,\ F\ensuremath{{}_{2}})} --- so
binding a role commutes with substitution of any filler. (ii) is the
statement that \texttt{F\ \ensuremath{\mapsto}\ R\ \ensuremath{\oplus}\ shift(F)} is a fixed function of F
parameterized only by R, applying with the same algebraic effect to any
filler, encountered or novel. (iii) is what defeats HRR and learned
embeddings and requires invertibility: because XOR is self-inverse and
shift is bijective, \texttt{Unbind(B,\ R)\ =\ shift\ensuremath{{}^{-}}\ensuremath{{}^{1}}(B\ \ensuremath{\oplus}\ R)\ =\ F}
recovers the filler exactly under noiseless conditions and approximately
under sparse-coding noise. Given \texttt{Bind(Stem,\ walk)}, querying
with \texttt{Stem} returns \texttt{walk}; the identical operation
returns \texttt{blicket} from \texttt{Bind(Stem,\ blicket)} for a verb
never before seen. Free generalization is built into the algebra rather
than approximated by a trained mapping --- the precise sense in which
Pillar 1 separates a symbol manipulator from a function approximator.

The distinction is worth stating in its sharpest form, because it is the
one that scale does not close. A trained approximator holds a record of
outcomes; an algebra holds an operation. Records can be extended, and
extending them is what scale does. Operations cannot be extended,
because there is nothing to extend --- they were already defined
everywhere. Asking for the past tense of a word never encountered is
asking what the system would do along a road it has not travelled, and a
representation that stores only where it has been cannot answer, however
much of the road it has covered. The difference is one of type, not of
size. Section 13 shows that this is the same observation Pearl makes
about the second rung of his ladder, arrived at from a different
direction.

\begin{center}\rule{0.5\linewidth}{0.5pt}\end{center}

\hypertarget{pillar-2-structured-representations-and-the-recursion-where-thagards-substrate-fails}{%
\section*{5. Pillar 2 --- structured representations, and the recursion
where Thagard's substrate
fails}\label{pillar-2-structured-representations-and-the-recursion-where-thagards-substrate-fails}}

\hypertarget{what-marcus-asked-and-what-the-engine-supplies}{%
\subsection*{5.1 What Marcus asked, and what the engine
supplies}\label{what-marcus-asked-and-what-the-engine-supplies}}

Pillar 2 asks that constituents in different structural positions yield
distinguishable representations, that each constituent be recoverable in
its role, and that recursion continue without exponential growth in
size. A bundle of role--filler bindings is the XOR sum
\texttt{Repr\ =\ \ensuremath{\oplus}\ensuremath{{}_{i}}\ (R\ensuremath{{}_{i}}\ \ensuremath{\oplus}\ shift(F\ensuremath{{}_{i}}))}. Distinguishability follows
from the non-commutativity of bind:
\texttt{Bind(Inner,\ Book)\ \ensuremath{\oplus}\ Bind(Outer,\ Table)} differs from
\texttt{Bind(Inner,\ Table)\ \ensuremath{\oplus}\ Bind(Outer,\ Book)} despite identical
role and filler multisets, recovering the book/table contrast at the
level of the bundled vector. Per-constituent recovery follows from
unbind: querying \texttt{Repr} with \texttt{R\ensuremath{{}_{j}}} returns
\texttt{F\ensuremath{{}_{j}}\ \ensuremath{\oplus}\ \ensuremath{\xi}\ensuremath{{}_{j}}}, where the crosstalk \texttt{\ensuremath{\xi}\ensuremath{{}_{j}}} from the other
bindings, by the quasi-orthogonal-diffusion property, scatters across
the address space near Hamming weight m/2 and is filtered by the
contractive majority-vote readout. This is exactly the recovery circular
convolution could not give: there the crosstalk accumulates \emph{as
bona fide noise in the same subspace as the signal}; here it is
scattered and filtered. A structurally scattered residual and an
algebraically accumulated one are not two amounts of one quantity ---
they are different objects, and no downstream cleverness converts the
second into the first. Fixed-dimensional recursion follows from
linearity --- a bundle is a hypervector in the same space as its
constituents and can be bound again as a filler --- so the five-level
embedding that costs on the order of 10\ensuremath{{}^{7}} nodes in tensor-product
calculus uses one fixed L-bit vector, bounded by collision saturation
rather than dimensional growth.

\hypertarget{the-recursion-the-thagard-viewpoint-foregrounds}{%
\subsection*{5.2 The recursion the Thagard viewpoint
foregrounds}\label{the-recursion-the-thagard-viewpoint-foregrounds}}

Here the reframing earns its place, because this is the single point at
which the \emph{reversibility} of the operation --- not merely its cost
--- changes what can be \emph{claimed}. Thagard's higher cognition is
deeply nested by construction: an emotion binds a self; the self is a
bound composite; an intention binds the emotion and the self again.
Circular convolution's quasi-inverse accumulates error at each level, so
SPA's own commitment to systematicity and productivity sits in tension
with a substrate that degrades with depth --- the seam identified in
\S{}2.2. XOR-binding removes the tension: because each level is exactly
invertible, an arbitrarily deep nesting decomposes without cumulative
loss in the noise-free limit. This is why VaCoAl is not merely a cheaper
substrate for Thagard's constructions but a \emph{faithful} one --- the
first on which his recursively bound self and emotions do not decay with
the depth his theory demands. The soap-opera chain of \emph{Amy loves
Billy, Billy loves Clara, \ldots, Henry loves Amy} is the small
diagnostic of the same property: eight role-asymmetric bindings whose
ironic closure is recovered by unbinding at fixed dimension, with
confidence \texttt{CR2\ =\ \ensuremath{\prod}\ CR1}, without retraining.

Two features of that small example are easy to pass over. The closure is
recovered by \emph{subtraction}: removing seven facts from the bundle
exactly leaves the eighth. Subtraction of a constituent is available only
where composition was exact; in a lossy scheme one can add facts to a
bundle but never cleanly remove one, so the eighth fact would have to be
inferred from what remains rather than read off it. And the confidence
attached to the answer is a \emph{product} of per-step margins, not a
single similarity score --- a statement about the route by which the
answer was reached rather than about the answer alone. Section 7 shows
what that quantity costs, and Section 13 shows what it buys.

A word on the two registers of evidence in this program, so their
pairing reads as method rather than mismatch. The small, closed chains
--- the book/table contrast, the eight-fact soap opera --- are treated
\emph{analytically}: their binding-multiplicity is low enough --- a
handful of bound pairs, a short bounded chain of a few hops --- that the
reversible algebra proves the recovery in closed form, and no
implementation is needed to establish it. The large, open structures ---
the 25.5-million-path scholarly genealogy on which the higher-level
claims rest --- cannot be traced in closed form and are therefore
established \emph{empirically}, in the extended PyVaCoAl realization
(\S{}3.0), where CR2 selection and the Rescue/Don't Care contrast are
measured rather than derived. This is a deliberate division of labour:
closed-form proof where the structure is small enough to admit it,
implemented demonstration where it is not. The two levels are
complementary evidence for one property --- exact reversible composition
--- not two inconsistent standards applied to one claim.

\begin{center}\rule{0.5\linewidth}{0.5pt}\end{center}

\hypertarget{pillar-3-individuals-versus-kinds-under-reversible-indexing}{%
\section*{6. Pillar 3 --- individuals versus kinds under reversible
indexing}\label{pillar-3-individuals-versus-kinds-under-reversible-indexing}}

Pillar 3 requires that the mind hold a persistent identifier per
individual, separate from the individual's current properties; bind the
individual to its kind without collapsing the two; and add new
individuals without retraining. VaCoAl supplies each under the same
algebra. Persistent identity is carried by an Entry Address (EA), a
distinct hypervector assigned at first encoding and quasi-orthogonal to
every other EA with overwhelming probability; the EA persists across
encounters while the bindings that link it to observed properties vary.
Kind is a binding, \texttt{Kinds(EA\_Felix)\ =\ Bind(ISA,\ Cat)}, while
properties are a separate bundle,
\texttt{Bind(HasColor,\ Orange)\ \ensuremath{\oplus}\ Bind(HasAge,\ Five)\ \ensuremath{\oplus}\ \ldots{}}; the two
live in one space and one algebra but are addressable separately because
\texttt{ISA} and \texttt{HasColor} are distinct role vectors, so
\emph{Felix is a cat} is operationally distinct from \emph{Felix is
orange and aged five}. Adding an individual is the trivial deposit of a
fresh EA from the quasi-orthogonal address space, disturbing no existing
representation --- the engineering counterpart of adult dentate
neurogenesis {[}13, 2{]} supplying fresh granule cells with new
connectivity. Crucially, kind and individual vectors are drawn from the
same space under the same algebra; the asymmetry between them is
relational (an individual's EA links to a kind through an ISA binding; a
kind links to its members through many individuals' kind bindings), not
a second representational format. Pillar 3 is met by \emph{indexing},
not new machinery --- a parsimony unavailable to the 2001 alternatives.

It is worth registering how much of Marcus's Chapter 5 worry this
dissolves and how much it does not. What it dissolves is the collapse:
two encounters with the same EA are the same individual by construction,
and two distinct EAs are distinct individuals however similar their
property bundles, because the identifier is not derived from the
properties. What it does not dissolve is the tracking problem ---
deciding, on encountering an orange cat, whether to retrieve
\texttt{EA\_Felix} or to mint a fresh one. That decision is a retrieval
question, answered by the readout with confidence CR1, and it is
fallible in exactly the way one would expect. The architecture
guarantees that individuals \emph{can} be kept distinct; it does not
guarantee that they always \emph{will} be, and no architecture that must
sometimes recognize things could.

\hypertarget{why-this-is-one-shot-and-continual-by-construction-not-by-training}{%
\subsection*{6.1 Why this is one-shot and continual by construction, not
by
training}\label{why-this-is-one-shot-and-continual-by-construction-not-by-training}}

The deposit of a fresh EA is worth dwelling on, because it makes
explicit a property that is definitional to this class of architecture
rather than an achievement to be optimized for. VaCoAl, like Kanerva's
Sparse Distributed Memory {[}32{]} from which it descends, is a
\emph{write-as-you-go} associative memory: a new individual, a new fact,
a new binding is stored by writing it into the content-addressable
memory in place, not by running gradient descent over a dataset. There
is no training loop to converge, no weights to update, no catastrophic
interference with prior representations --- a single presentation
suffices to store, and every subsequent write simply accumulates.
One-shot and few-shot acquisition are therefore not emergent
capabilities coaxed out of a trained model but \emph{structural
consequences} of the memory discipline itself; the same discipline makes
learning continual and lifelong by default, since the memory grows in
real time as encounters arrive rather than in an offline training phase
that must be frozen before deployment. This is the sense in which the
substrate learns the way SDM learns, and it is a property VaCoAl shares
with the broader family of hyperdimensional and vector-symbolic
architectures rather than one it must argue for. Where the earlier
pillars established that the algebra can \emph{represent} structure,
this observation establishes that it can \emph{acquire} structure
incrementally, at encounter time, without a separate learning regime ---
the precondition for an agent that must keep functioning while the world
keeps changing.

The content-addressable memory that supports this is also the organ of
analogical problem-solving. Retrieval in a CAM is cue-driven: a partial
or noisy probe returns the stored representation nearest to it, and ---
because binding is exactly invertible here --- the role structure of
what is retrieved is preserved rather than approximated. Confronted with
a novel situation, the system does not need a retrained model; it probes
its associative memory with the new situation as cue and recovers a
structurally related prior case, whose relations it can then map onto
the new one. In the terms of \S{}9.1, this cue-driven,
structure-preserving retrieval \emph{is} the mechanism of analogy, and
its fidelity is bounded by the fidelity of the CAM --- which is exactly
why exact reversible recall matters for it.

\begin{center}\rule{0.5\linewidth}{0.5pt}\end{center}

\hypertarget{no-failure-no-preference}{%
\section*{7. No failure, no preference: the Rescue-Rate phase
transition}\label{no-failure-no-preference}}

The preceding three sections established what the algebra can represent.
This section reports what the architecture had to give up in order to be
able to \emph{choose}, and it is the finding we would most like carried
out of this paper, because it does not depend on our substrate being
right about anything else.

\hypertarget{the-transition}{%
\subsection*{7.1 The transition}\label{the-transition}}

PyVaCoAl exposes the rescue circuit as a tunable parameter,
\texttt{RR\ \ensuremath{\in}\ {[}0,1{]}}. Set \texttt{RR\ =\ 1} and every block
collision is repaired from a rescue table built during the write phase.
The result is a system with no retrieval failures at all: within a given
frontier, its output is bit-identical to a plain hash dictionary,
which is to say it is exactly correct by the only external standard
available. And precisely because it is exactly correct, CR1 is pinned to
1.0 at every step, so CR2 = \ensuremath{\prod}\ensuremath{{}_{i}} CR1(i) is identically 1
for every candidate, and the ordering among reachable candidates falls
back on the host language's lexicographic default --- a criterion with no
semantic content whatever.

Set \texttt{RR\ =\ 0} at sufficient memory depth and collisions become
rare but not absent. CR1 now sits just below unity, varying slightly
from step to step according to how many blocks were lost; CR2 accumulates
those variations multiplicatively down a path; and that accumulated
quantity is the only thing distinguishing one reachable candidate from
another. Everything the architecture knows about the relative quality of
two routes it learned by failing, in small amounts, along them.

\begin{center}
\small
\begin{tabular}{@{}p{3.6cm}p{5.2cm}p{5.2cm}@{}}
\toprule
\textbf{Operating mode} & \textbf{Property gained} & \textbf{Property relinquished} \\
\midrule
Rescue (\texttt{RR\ =\ 1}) & Frontier-relative exact match; bit-identical agreement with a hash dictionary; zero collisions & CR2 discriminating power --- every candidate homogenized to CR2 = 1, ordering reverts to lexicographic \\
\addlinespace
Don't Care (\texttt{RR\ =\ 0}) & Path-dependent ranking by accumulated confidence; semantic selection & Zero-collision guarantee --- minute collisions tolerated by design \\
\bottomrule
\end{tabular}
\end{center}

\begin{center}\small\emph{Table 2. The two operating modes are the same circuit under one
parameter. The trade is not between a better and a worse setting but between two
different things a memory can be.}\end{center}

\hypertarget{the-general-form}{%
\subsection*{7.2 The general form}\label{the-general-form}}

It is easy to read Table 2 as a parameter trade-off. It is a constraint.

The information needed to prefer one retrieval over another is generated
only as a by-product of failing along it. Path quality, retrieval
confidence, and any answer to \emph{how far back should I go?} are all
differences in how badly things went; a system that never goes badly has
no differences to report. Perfect recall and zero discrimination are not
in tension here --- they are the same fact stated twice. This is why the
trade in Table 2 is not a defect of the present implementation to be
designed away in a later version: any architecture that repairs all of
its errors forfeits, by that act, the ability to rank what it retrieves.

The design consequence is that error tolerance should be exposed as a
control surface rather than minimized. VaCoAl makes the rescue circuit
switchable for exactly this reason: Rescue mode when frontier-relative
exactness is what is wanted, Don't Care mode when the question is which
of several reachable answers deserves to be believed. We are not aware
of another substrate in which the two regimes are the same circuit under
a single parameter, and we note the reflex the finding cuts against ---
that error suppression is an unqualified good, to be pushed as far as
the budget allows.

\hypertarget{why-this-matters-to-thagard}{%
\subsection*{7.3 Why this matters for the Thagard
reading}\label{why-this-matters-to-thagard}}

The point is not merely architectural, and this is where the Thagard
viewpoint again does work the Marcus reading alone could not. Thagard's
consciousness is a \emph{competition} among semantic pointers for a
limited broadcast: a mechanism whose entire function is to prefer one
candidate over others under a capacity constraint. A competition requires
a quantity to compete on. If every candidate arrives with identical
credentials, the winner is decided by whatever arbitrary tiebreak the
implementation happens to use, and calling that a competition flatters
it.

What \S{}7.1 shows is where such a quantity can come from in a
deterministic substrate: not from a separate scoring module bolted on,
but from the memory's own record of the difficulty it encountered. This
is a claim about our architecture, not about the brain, and we do not
extend it to a prediction about neural selection mechanisms. But it does
change the shape of the question one asks of Thagard's account. The
interesting question is no longer only \emph{what competes} but
\emph{what the competitors are graded on, and where that grade was
manufactured} --- and the answer, in at least one working system, is that
it was manufactured out of failure.

\begin{center}\rule{0.5\linewidth}{0.5pt}\end{center}

\hypertarget{convergent-evolution-not-biomimicry-why-the-reversible-layer-is-necessary}{%
\section*{8. Convergent evolution, not biomimicry: why the reversible
layer is
necessary}\label{convergent-evolution-not-biomimicry-why-the-reversible-layer-is-necessary}}

A complementary-layer proposal must answer one hard question: is the
missing layer actually \emph{needed}? The answer is not intuition but
convergence --- and the Thagard viewpoint is what makes the convergence
argument as strong as it is, because it adds a second independent
cognitive program to the tally.

\hypertarget{the-shared-function-reached-from-different-premises}{%
\subsection*{8.1 The shared function, reached from different
premises}\label{the-shared-function-reached-from-different-premises}}

We claim \emph{convergence}, not \emph{homology}, and the distinction is
load-bearing. Kanerva's blessing of dimensionality {[}33{]} --- that
random high-dimensional vectors are almost surely quasi-orthogonal,
permitting interference-free superposition --- is realized in the brain
through expansion recoding that manufactures genuine stochastic
independence, and in VaCoAl through the maximal-length sequence of a
primitive polynomial. The premises differ sharply: random projection
\emph{satisfies} the independence the theorem assumes, while VaCoAl's
deterministic diffusion does \emph{not} possess genuine independence and
instead \emph{acquires} quasi-orthogonality algebraically. Different
premises, same function is the textbook signature of convergent
evolution, as the feather and the membrane reach flight from different
materials. The ``quasi'' in quasi-orthogonality is therefore not a
weakness to be apologized for but the \emph{fingerprint} of convergence
--- the mark that VaCoAl arrived at the shared function by a route the
brain did not take.

\hypertarget{we-borrowed-the-problem-not-the-solution}{%
\subsection*{8.2 We borrowed the problem, not the
solution}\label{we-borrowed-the-problem-not-the-solution}}

Convergence carries a burden homology does not: it must defend
independence of origin, and VaCoAl was admittedly inspired by neural
theory. The defence is the discipline of \emph{functional induction}.
VaCoAl imitates none of the brain's physical structure --- no spikes, no
analog devices, no neurons --- but only the \emph{computational problem}
to be solved: finite-resource quasi-orthogonalization, reversible
composition, pattern completion. The solution, Galois-field algebra, is
drawn independently from mathematics with no biological content. We
borrowed the problem, not the solution --- as one might take the problem
of flight from birds while deriving the jet engine independently from
aerodynamics. This single distinction is what separates a convergence
claim from a biomimicry claim, and it is the deepest foundation stone of
the argument. Appendix A states the resulting relation precisely, as
\emph{structural flow equivalence} rather than mathematical isomorphism,
and marks where the weaker claim is all the evidence supports.

\hypertarget{the-root-constraint-and-the-two-program-tally}{%
\subsection*{8.3 The root constraint and the two-program
tally}\label{the-root-constraint-and-the-two-program-tally}}

The convergence is strongest stated as a root constraint with derived
consequences rather than a flat list. The root is finite-resource
quasi-orthogonalization: a bounded system that must superpose many
patterns on one substrate without catastrophic interference is driven
toward quasi-orthogonal codes; from this root, the binding problem
(composition without spurious conjunctions) and the capacity limits of
working memory follow as \emph{derived} constraints, not independent
ones. Now the tally the Thagard viewpoint completes: three independent
systems, under no common design lineage at the level of solution,
converge on the same reversible-compositional algebra --- Marcus's
symbol-manipulation pillars (from the mechanics of cognition), Thagard's
binding-based mind (from the architecture of emotion, consciousness, and
self), and the dentate gyrus--CA3 circuit (from biology, \S{}10). Two
cognitive-architecture programs derived from disjoint starting points,
plus a biological substrate, all requiring the same layer is the
evidence that the layer is not optional. It is what a mind that must
bind and later unbind is compelled to build.

There is a second, sharper reason the layer cannot be treated as
optional, and \S{}3.3 has already named it: it is not addable. An
architecture whose elementary binding is lossy does not acquire exact
decomposition by gaining a component, because exactness of inverse is a
property of the elementary operation and not of the apparatus stacked on
top of it. Compositionality is purchased with decomposability, and the
purchase is made at the level of the elementary operation or not at all.
A representation that cannot be exactly taken apart was never composed;
it was mixed. That is why the convergence tally reads as evidence about
a requirement rather than about a popular design choice --- there is no
version of these architectures that has the capability without having
the algebra.

\begin{center}\rule{0.5\linewidth}{0.5pt}\end{center}

\hypertarget{thagards-higher-cognition-on-a-reversible-substrate-as-witness-not-implementation}{%
\section*{9. Thagard's higher cognition on a reversible substrate --- as
witness, not
implementation}\label{thagards-higher-cognition-on-a-reversible-substrate-as-witness-not-implementation}}

We treat Thagard's boldest chapters --- emotion, consciousness, self ---
with deliberate restraint, because their role in this paper is
corroborative, not foundational, and over-reading them would forfeit the
discipline of \S{}8.

Read as a theory of mind, Thagard's account is not ours to adjudicate
here, and we do not implement consciousness. Read as \emph{evidence}, it
is directly to our purpose. His emotion --- a binding of situation,
physiology, appraisal, and self --- is a role-filler bundle of low
binding-multiplicity, recovered cleanly on our substrate at high CR:
exactly the case where exact unbinding is straightforwardly superior,
and, as \S{}1.3 argued, exactly the case in which the difference between
recovering the whole and recovering its constituents is the difference
between being moved and being informed. His self --- a recursive binding
of representations about oneself, with self-changing operations that are
literal re-bindings --- is where
\S{}5.2's depth-exact reversibility pays its clearest dividend, and where
auditability (the ability to decompose a self-representation and inspect
its provenance) is a capability rather than an ornament. His
consciousness --- a competition among semantic pointers for a limited
broadcast --- is mirrored not by our reversibility but by the
CR-weighted majority vote and Don't Care lateral inhibition that let one
candidate emerge while mismatched blocks diffuse into background --- and
this is a \emph{PyVaCoAl-demonstrated} behaviour, specific to the
\texttt{RR\ =\ 0}, sufficient-depth regime in which CR2 retains
discriminating power (\S{}7), not a property of the bare algebra; we note the
correspondence and rest no weight on it, because the claim we defend is
only the modest, checkable one --- that the same \emph{functional
organization} (role-filler binding, recursive nesting, competitive
selection under capacity limits) recurs across substrates --- and not
that our architecture is conscious. The hard problem is untouched by the
choice of binding operation, in our substrate as in Thagard's; swapping
convolution for XOR changes fidelity and cost, not the explanatory gap,
and we flag the boundary rather than blur it.

What the Thagard viewpoint yields, precisely, is this: his
higher-cognitive constructions become an independently evolved instance
of the reversible-compositional layer, realized on tissue; that our
deterministic substrate converges on the same organization is the
corroboration that the layer is substrate-independent, and hence
necessary; and where his substrate strained under recursion, ours does
not.

\hypertarget{analogy-as-reversible-associative-recall}{%
\subsection*{9.1 Analogy as reversible associative
recall}\label{analogy-as-reversible-associative-recall}}

There is one further construction of Thagard's on which the substrate
bears directly, and it is the one his broader body of work makes
central: analogy. In Thagard's account analogy is not a peripheral trick
but a core operation of cognition, and it is \emph{structure-mapping}
--- the transfer of a system of relations, not a list of surface
attributes, from a source case to a target --- governed by
systematicity, the principle that it is the higher-order relational
structure, and the causal relations among relations, that ought to be
carried across. Crucially, in a memory-based reading analogy is executed
\emph{by the associative recall itself}: to probe memory with a present
situation and retrieve a structurally related prior case is already to
perform the mapping, provided the retrieval preserves the relational
structure of what it returns. This is precisely where the choice of
binding operation ceases to be an implementation detail. A retrieval
whose unbinding is only approximate --- circular convolution's
quasi-inverse --- degrades the relations it carries at every hop, so
that systematicity erodes exactly as the mapping deepens; the
higher-order structure that analogy is supposed to preserve is the first
thing an accumulating quasi-inverse loses. Exact reversible unbinding
carries the relational structure across without loss, so that the
recalled source arrives with its role bindings intact and available to
be mapped. The content-addressable memory is therefore not merely
\emph{where} analogies are stored but the organ that \emph{executes}
them, and the accuracy and coverage of that memory set the accuracy and
coverage of the analogies it can draw --- the barometer, in the plainest
sense, of analogical competence. This is the analogical counterpart of
\S{}5.2's recursion argument: the property that made deep nesting decompose
without loss is the same property that lets structure-mapping run
without eroding systematicity, and it is the reason a reversible
associative memory, rather than a lossy one, is the substrate on which
Thagard's analogy can actually be carried out. We note, keeping the
discipline of \S{}8, that this establishes a \emph{mechanism}, not a claim
to have reproduced human analogical judgment; the empirical measurement
of analogical transfer on the substrate remains, like the other
higher-level claims, a PyVaCoAl-level demonstration to be reported
rather than an architectural entitlement.

\begin{center}\rule{0.5\linewidth}{0.5pt}\end{center}

\hypertarget{neural-implementation-dgca3-as-the-biological-engine-in-brief}{%
\section*{10. Neural implementation: DG--CA3 as the biological engine (in
brief)}\label{neural-implementation-dgca3-as-the-biological-engine-in-brief}}

The companion Perspective {[}6{]} argues that the dentate gyrus--CA3
circuit instantiates the same engine through five biophysical
homologues; we summarize only what \S{}8's convergence tally and Marcus's
open questions require. \textbf{GF(2) register state} --- sparse,
theta--gamma-clocked granule-cell activation {[}7, 15, 21{]} realizes
the clocked sparse binary register, answering Marcus's Open Question C.
\textbf{XOR primitive} --- dendritic action potentials in human cortical
pyramidal neurons implement XOR-like input--output transformations
{[}10{]}, with the hippocampal pyramidal family {[}3{]} supplying the
same nonlinear integration {[}14{]}. \textbf{Deterministic propagation}
--- mossy-fiber detonator transmission {[}12, 28, 4{]} reliably
propagates single granule-cell activations, eliminating random
projection's bounded-failure mode at the substrate rather than
statistically. \textbf{Structural specification} --- the sparse,
directed, developmentally programmed mossy-fiber projection {[}1, 8, 23,
29{]} defines a specific operation per cell, the biological correlate of
selecting a generator polynomial \texttt{Gb(x)}, and the answer to
Marcus's Open Question B: developmentally specified targeting is exactly
the innate, non-blueprint microcircuitry that cascading master-control
genes can produce {[}9, 27{]}. \textbf{Contractive readout} --- CA3 recurrent
attractor dynamics {[}11, 24{]} collect block-wise outputs, the
biological analogue of majority voting. The honest claim is not that
these answers are final --- bit-exact reversibility holds in silicon but
only approximately under biological noise, and the
developmental-program-to-polynomial correspondence is structural rather
than proven by functional-equivalence theorems --- but that Marcus's
open questions now have candidate, testable answers at the level of
substrate.

We are careful about the kind of claim this is, and Appendix A states it
in full. It is not an argument that the engine mimics the hippocampus,
nor that the hippocampus implements a shift register. It is an argument
of convergent design: two substrates, one carbon and one silicon, faced
the same computational problem --- orthogonalize sparse inputs,
propagate them without loss of identity, contract many local verdicts
into one --- and arrived at structurally similar solutions because the
problem admits few.

\begin{center}\rule{0.5\linewidth}{0.5pt}\end{center}

\hypertarget{what-this-is-that-the-2001-alternatives-and-the-2024-default-were-not}{%
\section*{11. What this is, that the 2001 alternatives --- and the 2024
default --- were
not}\label{what-this-is-that-the-2001-alternatives-and-the-2024-default-were-not}}

Table 1 states the trades; this section says what each one costs.
\textbf{Tensor products} {[}26{]} occupy the same algebraic role
(non-commutative binding with exact unbind) but at a dimension that is
the product of constituent dimensions, so a five-level embedding needs
on the order of 10\ensuremath{{}^{7}} nodes; VaCoAl achieves the same at fixed dimension,
bounded by collision saturation scaling logarithmically in block count.
The tensor product is not accused of getting the algebra wrong: it gets
the algebra right and cannot afford it, and VaCoAl keeps the same two
properties while changing the bill. \textbf{Holographic reduced
representations} {[}22{]} achieve
fixed-dimensional recursion but replace exact unbind with a
quasi-inverse whose noise accumulates with bundle size and depth, ruling
out the bit-exact compositional inspection Pillar 2 wanted --- and, as
\S{}2.2 stressed, the exact substrate on which Thagard's recursive self and
emotions decay. VaCoAl keeps HRR's fixed dimension and restores
exactness: unbind is the algebraic inverse of bind, not its
approximation, and crosstalk is scattered across the address space
rather than accumulated in the signal subspace. HRR is the closest of
the alternatives and the most instructive failure, which is why \S{}8.3 is
written about it rather than about the others. \textbf{Temporal
synchrony and asynchrony} {[}25, 16{]} face a combinatorial-precision
problem as propositions multiply; VaCoAl replaces temporal precision
with algebraic diffusion, and the binding asymmetry comes from shift
rather than firing time, with discrimination exponentially tight in m
rather than linearly tight in timing --- a change in how discrimination
scales, not an improvement in how carefully it is done.

\textbf{The 2024 default --- learned dense embeddings --- is the case
the Thagard viewpoint makes sharpest, and it deserves its own statement
as an impossibility rather than a weakness.} Composition in a
transformer is attention-weighted superposition followed by nonlinear
transformation: a continuous, gradient-optimized mixture on which no
algebraic inverse is defined, so recovery of a constituent is at best a
learned approximation whose error compounds with composition depth. The
decisive point is that this is a property of the representational
\emph{class}, not the model \emph{instance}. Enlarging the parameter
count enriches the mixture and sharpens the approximate read-out; it
does not install an exact inverse, because exactness of inverse is an
algebraic property a lossy continuous superposition does not have and
cannot be trained into having. One does not make a projection invertible
by widening it. Hence the capacity that strict multi-hop relational
reasoning requires --- compose, then recover \emph{exactly}, at depth
--- is not a frontier scaling crosses; it is orthogonal to the axis
along which scaling operates. This is why the substrate is
\emph{complementary}, not competitive: the LLM is not deficient at what
it is for; it is being asked, in multi-hop relational tasks, to do what
its representation structurally cannot, and VaCoAl does it on a
different axis that composes with the LLM's own.

\begin{center}\rule{0.5\linewidth}{0.5pt}\end{center}

\hypertarget{resolution-as-part-of-the-answer}{%
\section*{12. Resolution as part of the
answer}\label{resolution-as-part-of-the-answer}}

The parameters of \S{}3.1 have an implication for output format that we
state separately, because it does not follow from any single
configuration and is easy to file as a limitation.

The engine has two free parameters, the block count N and the memory
depth per block 2\ensuremath{{}^{m}}, and under a fixed total capacity they trade
against each other. Raising N buys orthogonal resolution --- more
independent local verdicts, finer discrimination among structures --- at
the cost of a shallower per-block address space and therefore more
collisions. Raising m buys collision headroom at the cost of resolution.
Neither setting is correct. Each yields a representation that is
internally coherent, decomposable to a characteristic depth, and honest
about its own limits through CR1 and CR2; and, as the empirical record
of the companion engineering paper {[}5{]} documents, the structures
that survive to the top of a deep composition are \emph{not the same}
across settings. Different resolutions surface different macroscopic
structures, each internally coherent, from one corpus.

The natural output of such an architecture is therefore not a single
representation but a family indexed by resolution --- closer to a
spectrum than to an answer. Reporting that one structure dominates at
one resolution and another at a second, together with the parameter that
selects between them, is strictly more informative than reporting either
alone, and it is a report a system whose output type is one plausible
continuation cannot make. The resolution parameter is not noise to be
tuned away before publication; it is part of what was found.

This bears on the Thagard reading in a specific way, and we put it no
more strongly than the evidence allows. A mind that must sometimes see a
situation coarsely and sometimes finely is a mind for which resolution
is a control surface rather than a constant --- and an architecture in
which the same corpus yields different coherent readings at different
resolutions is at least the right \emph{kind} of architecture for that.
We do not claim to have explained anything about human construal. We
observe that the alternative --- an architecture that produces one
reading and cannot report the others --- has no place to put the
phenomenon at all.

\begin{center}\rule{0.5\linewidth}{0.5pt}\end{center}

\hypertarget{outlook-from-the-algebraic-mind-to-counterfactual-reasoning}{%
\section*{13. Outlook: from the Algebraic Mind to counterfactual
reasoning}\label{outlook-from-the-algebraic-mind-to-counterfactual-reasoning}}

Section 1.3 argued that introspection is a decomposition problem and
Section 4 that free generalization is a question about a road not
travelled. This section closes both loops.

Marcus's program targets symbol manipulation; Pearl {[}19, 20{]}
classifies cognition by causal query --- rung 1 (association,
\texttt{P(Y\textbar{}X)}), rung 2 (intervention,
\texttt{P(Y\textbar{}do(X))}), rung 3 (counterfactuals,
\texttt{P(Y\ensuremath{{}_{x}}\textbar{}X\ensuremath{{}^{\prime}},Y\ensuremath{{}^{\prime}})}) --- and shows that rungs 2--3 are
unreachable by pure pattern association. The engine's commitments map
onto the ladder. Rung 1 is associative retrieval by block-wise voting
with confidence CR1. Rung 2 is the surgical modification of one binding:
unbind \texttt{R\ensuremath{{}_{j}}} to recover \texttt{F\ensuremath{{}_{j}}}, substitute a counterfactual
\texttt{F\ensuremath{{}^{\prime}}\ensuremath{{}_{j}}}, and rebind --- a sequence of XOR-and-shift operations that
touches only the \texttt{R\ensuremath{{}_{j}}} binding, where convolution's lossy unbind
would accumulate noise from all others. The word \emph{surgical} is
doing real work: an intervention that perturbs the bindings it was
supposed to hold fixed is not a less precise intervention but a
different operation. Rung 3 requires simultaneous
factual and counterfactual worlds; the companion paper argues that the
two-orthogonalizer hippocampal architecture --- a scaffold regime for
the factual, mossy-fiber writes for the counterfactual --- supplies the
parallel non-interfering substrate, with CR2 path traces estimating
\texttt{P(Y\ensuremath{{}_{x}}\textbar{}X\ensuremath{{}^{\prime}},Y\ensuremath{{}^{\prime}})}. This extends Marcus's program rather than
belonging to it: he did not target counterfactuals in 2001, but the
substrate his pillars require turns out to also support the rung-3
capacity pattern associators cannot --- the Algebraic Mind as the
substrate-level prerequisite for the counterfactual reasoning Pearl
identifies, and, read through Thagard, for the self-projection into
contrary-to-fact scenarios that imagination and planning require.

Seen this way the three demands of this paper are one demand. Marcus
asks that a rule extend to an instance never encountered. Thagard's mind
asks what its own emotion was made of, and what it would have felt had
the appraisal been different. Pearl asks what would have happened along
a road not taken. Each is unanswerable by any representation that
records outcomes rather than holding operations, and each is answered,
when it is answered, by the ability to unbind at a site, substitute, and
rebind without disturbing anything else. That the same three lines of
XOR-and-shift serve all three is the reason we think the requirement is
structural rather than disciplinary.

A second, more concrete outlook concerns the VaCoAl/CASRAM boundary
flagged throughout \S{}3.0 and \S{}14. The empirical record in this paper is
exclusively a PyVaCoAl record; the architectural low-power,
nanosecond-scale claim is exclusively a VaCoAl/CASRAM claim; and nothing
here closes that gap. What can be reported is that the gap is now a
measurement to be taken rather than a design to be built: CASRAM exists
as a commercial SRAM/DRAM-CAM realization of the architecture (Shuhari
System), and an author--Shuhari System collaboration aimed at
benchmarking PyVaCoAl's demonstrated million-dimensional, multi-hop
reasoning directly on CASRAM's physical substrate is beginning. We
report this as a concrete next step, not as a result of this paper:
until that benchmark is run and published, the reader should continue to
treat every speed and power figure attached to VaCoAl as architectural,
and every reasoning-capability figure attached to PyVaCoAl as the only
one this paper actually measured.

\begin{center}\rule{0.5\linewidth}{0.5pt}\end{center}

\hypertarget{reservations-and-testable-predictions}{%
\section*{14. Reservations and testable
predictions}\label{reservations-and-testable-predictions}}

\textbf{Reservations.} (i) Bit-exact reversibility holds in silicon but
only approximately under biological noise --- potentiation and
depression are stochastic, recall under interference is graded, and the
population-level reversibility attributed to DG--CA3 holds only in a
statistical sense requiring sparse coding; HRR faces the same gap
relative to its idealized form, and the engine reading inherits it. The
reservation bounds claims about exact biological realization but not the
architectural correspondence with Marcus's pillars --- the algebra is
the algebra. It also interacts with \S{}8.3 in a way we should not paper
over: if biological binding is only approximately reversible, then the
necessity argument implies that biological compositional inspection is
correspondingly bounded, which is an empirical prediction and not
obviously a comfortable one. (ii) The molecular correlate of selecting \texttt{Gb(x)} is
unidentified; mossy-fiber specificity is developmentally programmed, but
its functional equivalence to choosing a primitive generator is
conjectural. (iii) The treelet-as-GF(2)-register-set reading is a
parsimony argument, not a necessity proof; Marcus's treelet is general
enough to admit other algebras, and we argue only that GF(2) is the most
parsimonious choice satisfying reversibility, non-commutativity, and
free generalization under one primitive. (iv) Most importantly for
scope: nothing here shows that our mechanism explains Thagard's
phenomena \emph{better} than his --- the necessity argument is
convergence, not explanatory victory, and the coherence comparison that
a genuine theory-replacement would demand is deliberately not attempted.
(v) The two-level distinction of \S{}3.0 bounds every capability claim
above. The reversible-binding correspondence with Marcus's pillars and
the recursion argument through Thagard rest on the shared algebraic core
and hold at the architectural level. But everything \emph{demonstrated}
--- the multi-hop relational tracing, the CR2 semantic selection and its
STDP-like Don't Care pruning, the Rescue/Don't Care mode contrast, the
million-dimensional scale --- is a property of \textbf{PyVaCoAl}, the
extended software realization on DRAM, and inherits its RR/memory-depth
phase transition: at \texttt{RR\ =\ 1} the system reduces to a hash
dictionary and the CR2 discrimination the higher-level reading relies on
vanishes, so the semantic-selection claims are claims about the
\texttt{RR\ =\ 0}, sufficient-depth regime specifically. The
\textbf{VaCoAl} in-memory speed and 1/N-activation low-power claims are
architectural and, within this paper, unbenchmarked --- a distinction we
hold apart from \emph{unbuilt}: the architecture is physically realized
as CASRAM (Shuhari System, \S{}3.0, \S{}13), so the open question is a pending
measurement on existing hardware, not an open question of buildability.
PyVaCoAl's measured performance is a lower bound on, not a demonstration
of, the CASRAM figures. Readers should not transfer a demonstrated
PyVaCoAl result to the hardware realization, nor a CASRAM-realized
VaCoAl property to the empirical record, without noting which level
carries it. (vi) The general claim of \S{}7.2 --- that ranking is purchased
with failure --- is stated at the level of this architecture and argued,
not proved, to hold beyond it; the next paragraph says how it could be
refuted.

\textbf{Testable predictions.} \emph{Architectural:} Marcus's Chapter 3
and 5 case studies --- inflection generalization to pseudo-verbs,
soap-opera role-filler tracking, object-permanence individual/kind
distinction --- should be solvable on the architecture with capacity
scaling logarithmically in block count, confidence given by CR1/CR2, and
failure modes predictable from collision saturation rather than training
coverage; direct implementation is a concrete test, and we state plainly
that it has not yet been run. \emph{Biological:}
the companion paper's predictions --- multiplicative decay in multi-hop
replay, absence of the random-projection bounded-failure mode at minimal
input perturbations, primate-versus-rodent per-cell pattern-separation
efficiency, encoding/retrieval lesion dissociation --- constrain the
biological reading independently. \emph{Cognitive, via the Thagard
viewpoint:} if recursive self- and emotion-binding rest on an exactly
reversible substrate, then tasks requiring deep recursive
self-representation should degrade with depth far more gently than a
convolutional substrate predicts, and the observed degradation should
follow the CR2 multiplicative curve rather than the accumulating-noise
curve of a quasi-inverse --- a dissociation that distinguishes the
engine reading from an SPA baseline on Thagard's own constructions.
\emph{Refutations we would accept:} an architecture that supports
bit-exact compositional inspection on top of a demonstrably lossy
elementary binding operation would refute \S{}8.3; a retrieval system that
repairs every failure and nevertheless ranks its reachable candidates on
a semantically contentful criterion generated internally, without an
external scoring module, would refute \S{}7.2. We do not think either
exists, but the burden is ours and this is where it should be
discharged.

\begin{center}\rule{0.5\linewidth}{0.5pt}\end{center}

\hypertarget{conclusion}{%
\section*{15. Conclusion}\label{conclusion}}

\emph{The Algebraic Mind} closed by asking how its three pillars are
implemented in neural hardware; \emph{Brain--Mind} answered a larger
question --- what binding, once available, can build --- and rested the
answer on a substrate that strains under its own ambition. Twenty-five
years and one paradigm of language models later, the substrate both
programs called for is available: an algebro-deterministic architecture
around a single primitive, XOR-and-shift over GF(2), that supports
reversible variable binding (Pillar 1), non-commutative compositional
bundling at fixed dimension (Pillar 2), and individual/kind separation
under the same algebra (Pillar 3), and that removes the
depth-degradation which afflicts convolution precisely where Thagard's
recursive self and emotions are deepest. That two independent cognitive
programs and one biological circuit converge on this same
reversible-compositional layer is our evidence --- by convergent
evolution, not biomimicry; we borrowed the problem, not the solution ---
that the layer is a necessity rather than a preference. We do not claim
to be the mind, to be conscious, or to surpass the large language model.
We claim to supply the reversible, auditable, multi-hop relational
reasoning that statistical embeddings structurally lack: an orthogonal
axis that composes with linguistic fluency into a stack neither
completes alone.

Two consequences deserve to be carried out of this paper separately from
the architecture that produced them.

The first is that compositionality is purchased with decomposability,
and the purchase is made at the elementary operation or not at all. A
binding that can only be approximately undone does not yield a slightly
degraded constituent; it yields a different kind of object --- a
neighbourhood rather than a part --- and no downstream cleverness
recovers the difference. This is why Marcus's three pillars turn out not
to be three requirements, and why the layer cannot be added to an
architecture that lacks it: there is no version of a lossy substrate
that has the capability without having the algebra. We expect the
constraint to hold well beyond the present substrate, and it cuts
against the reflex of treating approximation as a quality dial that can
be turned back up later.

The second is that the ability to rank what one retrieves is purchased
with the ability to fail at retrieving it. Rescue mode and Don't Care
mode are the same circuit under one parameter, and moving from the
second to the first buys bitwise agreement with a hash dictionary at the
cost of every candidate becoming indistinguishable from every other.
Path quality is a difference in how badly things went; a system that
never goes badly has no differences to report. For an architecture whose
higher-cognitive reading rests on \emph{competition} among candidates,
this is not a footnote: it says where the quantity competed on has to
come from.

Both properties are, in the end, the same property seen from two sides:
a representation that can be taken apart can be put back together
differently, and a memory that keeps a record of its own difficulties
can compare the roads it did not take. The Algebraic Mind, read through
\emph{Brain--Mind}, has a concrete substrate --- and it is the layer the
statistical mind was missing.

\begin{center}\rule{0.5\linewidth}{0.5pt}\end{center}

\hypertarget{appendix-a}{%
\section*{Appendix A. Structural flow equivalence, not mathematical
isomorphism}\label{appendix-a}}

Throughout this paper, and more extensively in the companion Perspective
{[}6{]}, we assert correspondences between an algebra realized in
silicon and mechanisms observed in tissue. It is worth setting out, in
one place, what kind of claim that is, because the available vocabulary
invites overstatement.

\textbf{What is not claimed.} We do not claim mathematical isomorphism.
An isomorphism demands a strict bidirectional structure-preserving map,
and no such map is exhibited here or, to our knowledge, exhibitable on
present evidence. Granule-cell ensembles are not elements of a GF(2)
vector space; mossy-fiber targeting is not known to be equivalent to the
selection of a primitive generator; dendritic nonlinearity is
XOR-\emph{like} in its input--output signature and is not an XOR gate.
Each of these gaps is real, and none is closed by \S{}10.

\textbf{What is claimed.} We claim \emph{structural flow equivalence}: a
correspondence in the computational flow ``division \ensuremath{\rightarrow} local
processing \ensuremath{\rightarrow} integration,'' together with a correspondence in what
each stage must accomplish for the whole to work. The dentate gyrus
divides an input across many sparsely active units; each unit performs a
local, structurally specified, nonlinear transformation; CA3 contracts
the resulting local verdicts into one coherent reading. The engine
divides a hypervector across blocks; each block performs a local,
polynomial-specified, reversible diffusion; majority voting contracts
the block outputs into one reading with confidence CR1. The two
descriptions are compatible at the level of flow and of function, and
incompatible at the level of substrate; the claim covers the first two.

\textbf{Why this is the right framing.} The relationship is the one
biology calls convergent evolution, and \S{}8.2 states the discipline that
earns the term: we borrowed the problem, not the solution. Read this
way, the correspondence carries information without carrying a claim of
identity --- it tells us that the constraints are tight, which is
exactly what one wants to know when asking whether a substrate could
have been otherwise. Read as mimicry, the same correspondence would be a
weaker and more easily refuted claim, since the engine plainly mimics
nothing: it was designed for speed, power, and cost on commodity
silicon, and the convergence was noticed afterwards.

\textbf{What would strengthen it.} Any of the following: a demonstration
that developmental specification of mossy-fiber targets selects
transformations with the maximal-period property that primitive
polynomials have; a measurement of population-level reversibility in
DG--CA3 under sparse coding that meets a stated numerical bound; or a
functional-equivalence theorem relating CA3 attractor readout to
block-wise majority voting under a specified noise model. Each is a
research programme rather than an experiment, and we state them so that
the present claim can be seen for the modest thing it is.

\textbf{Acknowledgments.} We thank Gary Marcus for the framework and the
clarity with which the 2001 program was stated, and we read Paul
Thagard's \emph{Brain--Mind} not as a target to be displaced but as the
account whose ambition told us what the substrate is for. The open
questions that motivated this work were articulated explicitly by both;
our contribution is to offer one candidate substrate against which their
programs can now be compared. We also note, for transparency, the
collaboration now beginning with Shuhari System to benchmark PyVaCoAl's
demonstrated reasoning directly on the CASRAM hardware realization of
VaCoAl (\S{}13); none of that benchmark's results are reported in this
paper.

\begin{center}\rule{0.5\linewidth}{0.5pt}\end{center}

\hypertarget{references}{%
\section*{References}\label{references}}

{[}1{]} Acsády, L., Kamondi, A., Sík, A., Freund, T., \& Buzsáki, G.
(1998). GABAergic cells are the major postsynaptic targets of mossy
fibers in the rat hippocampus. \emph{Journal of Neuroscience}, 18(9),
3386--3403.

{[}2{]} Akers, K. G., Martinez-Canabal, A., Restivo, L., et al.~(2014).
Hippocampal neurogenesis regulates forgetting during adulthood and
infancy. \emph{Science}, 344(6184), 598--602.

{[}3{]} Benavides-Piccione, R., Regalado-Reyes, M., Fernaud-Espinosa,
I., et al.~(2020). Differential structure of hippocampal CA1 pyramidal
neurons in the human and mouse. \emph{Cerebral Cortex}, 30(2), 730--752.

{[}4{]} Chamberland, S., Timofeeva, Y., Evstratova, A., Volynski, K., \&
Tóth, K. (2018). Action potential counting at giant mossy fiber
terminals gates information transfer in the hippocampus.
\emph{Proceedings of the National Academy of Sciences}, 115(28),
7434--7439.

{[}5{]} Chuma, H., Otsuka, K., \& Sato, Y. (2026a). Beyond LLMs, sparse
distributed memory, and neuromorphics: A hyper-dimensional SRAM-CAM
``VaCoAl'' for ultra-high speed, ultra-low power, and low cost.
\emph{arXiv preprint} arXiv:2604.11665.

{[}6{]} Chuma, H., Otsuka, K., \& Sato, Y. (2026b). Bridging silicon and
the hippocampus: Algebro-deterministic memory ``VaCoAl'' as a substrate
for Vector-HaSH and TEM. \emph{arXiv preprint} arXiv:2605.15652.

{[}7{]} Diamantaki, M., Frey, M., Berens, P., Preston-Ferrer, P., \&
Burgalossi, A. (2016). Sparse activity of identified dentate granule
cells during spatial exploration. \emph{eLife}, 5, e20252.

{[}8{]} Galimberti, I., Gogolla, N., Alberi, S., Santos, A. F., Muller,
D., \& Caroni, P. (2006). Long-term rearrangements of hippocampal mossy
fiber terminal connectivity in the adult regulated by experience.
\emph{Neuron}, 50(5), 749--763.

{[}9{]} Gerhart, J., \& Kirschner, M. (1997). \emph{Cells, Embryos, and
Evolution}. Oxford: Blackwell Science.

{[}10{]} Gidon, A., Zolnik, T. A., Fidzinski, P., et al.~(2020).
Dendritic action potentials and computation in human layer 2/3 cortical
neurons. \emph{Science}, 367(6473), 83--87.

{[}11{]} Guzman, S. J., Schlögl, A., Frotscher, M., \& Jonas, P. (2016).
Synaptic mechanisms of pattern completion in the hippocampal CA3
network. \emph{Science}, 353(6304), 1117--1123.

{[}12{]} Henze, D. A., Wittner, L., \& Buzsáki, G. (2002). Single
granule cells reliably discharge targets in the hippocampal CA3 network
in vivo. \emph{Nature Neuroscience}, 5(8), 790--795.

{[}13{]} Kempermann, G., Gage, F. H., Aigner, L., et al.~(2018). Human
adult neurogenesis: Evidence and remaining questions. \emph{Cell Stem
Cell}, 23(1), 25--30.

{[}14{]} Krueppel, R., Remy, S., \& Beck, H. (2011). Dendritic
integration in hippocampal dentate granule cells. \emph{Neuron}, 71(3),
512--528.

{[}15{]} Leutgeb, J. K., Leutgeb, S., Moser, M. B., \& Moser, E. I.
(2007). Pattern separation in the dentate gyrus and CA3 of the
hippocampus. \emph{Science}, 315(5814), 961--966.

{[}16{]} Love, B. C. (1999). Utilizing time: Asynchronous binding.
\emph{Advances in Neural Information Processing Systems}, 11.

{[}17{]} Marcus, G. F. (2001). \emph{The Algebraic Mind: Integrating
Connectionism and Cognitive Science}. Cambridge, MA: MIT Press.

{[}18{]} Mullally, S. L., \& Maguire, E. A. (2014). Memory, imagination,
and predicting the future: A common brain mechanism? \emph{The
Neuroscientist}, 20(3), 220--234.

{[}19{]} Pearl, J. (2009). \emph{Causality: Models, Reasoning, and
Inference} (2nd ed.). Cambridge: Cambridge University Press.

{[}20{]} Pearl, J., \& Mackenzie, D. (2018). \emph{The Book of Why: The
New Science of Cause and Effect}. New York: Basic Books.

{[}21{]} Pernía-Andrade, A. J., \& Jonas, P. (2014).
Theta-gamma-modulated synaptic currents in hippocampal granule cells in
vivo define a mechanism for network oscillations. \emph{Neuron}, 81(1),
140--152.

{[}22{]} Plate, T. A. (1995). Holographic reduced representations.
\emph{IEEE Transactions on Neural Networks}, 6(3), 623--641.

{[}23{]} Rollenhagen, A., \& Lübke, J. H. R. (2010). The mossy fiber
bouton: the ``common'' or the ``unique'' synapse? \emph{Frontiers in
Synaptic Neuroscience}, 2, 2.

{[}24{]} Rolls, E. T. (2023). The hippocampus, ventromedial prefrontal
cortex, and episodic and semantic memory. \emph{Progress in
Neurobiology}, 217, 102334.

{[}25{]} Shastri, L., \& Ajjanagadde, V. (1993). From simple
associations to systematic reasoning: A connectionist representation of
rules, variables, and dynamic bindings using temporal synchrony.
\emph{Behavioral and Brain Sciences}, 16(3), 417--451.

{[}26{]} Smolensky, P. (1990). Tensor product variable binding and the
representation of symbolic structures in connectionist systems.
\emph{Artificial Intelligence}, 46(1--2), 159--216.

{[}27{]} Südhof, T. C. (2008). Neuroligins and neurexins link synaptic
function to cognitive disease. \emph{Nature}, 455(7215), 903--911.

{[}28{]} Vyleta, N. P., Borges-Merjane, C., \& Jonas, P. (2016).
Plasticity-dependent, full detonation at hippocampal mossy fiber--CA3
pyramidal neuron synapses. \emph{eLife}, 5, e17977.

{[}29{]} Wilke, S. A., Antonios, J. K., Bushong, E. A., et al.~(2013).
Deconstructing complexity: Serial block-face electron microscopic
analysis of the hippocampal mossy fiber synapse. \emph{Journal of
Neuroscience}, 33(2), 507--522.

{[}30{]} Eliasmith, C. (2013). \emph{How to Build a Brain: A Neural
Architecture for Biological Cognition}. Oxford: Oxford University Press.

{[}31{]} Thagard, P. (2019). \emph{Brain--Mind: From Neurons to
Consciousness and Creativity}. Cambridge: Cambridge University Press.

{[}32{]} Kanerva, P. (1988). \emph{Sparse Distributed Memory}.
Cambridge, MA: MIT Press.

{[}33{]} Kanerva, P. (2009). Hyperdimensional computing: An introduction
to computing in distributed representation with high-dimensional random
vectors. \emph{Cognitive Computation}, 1(2), 139--159.

{[}34{]} Shuhari System (2026). CASRAM product page.
\url{https://shuharisystem.com/?page_id=1336} (accessed July 2026). Cited as
the existing physical realization of the VaCoAl architecture {[}5{]}; no
benchmark data from this product is reported in this paper (\S{}3.0, \S{}11,
\S{}12).

\emph{References {[}1{]}--{[}29{]} reproduce, with light restyling, the
bibliography of the companion technical paper (arXiv:2605.21379v2), to
which this draft's Marcus-pillar correspondence and DG--CA3 reading are
indebted; {[}30{]}--{[}34{]} are added in this revision for the sources
--- Eliasmith, Thagard, Kanerva, and CASRAM --- that the Thagard
viewpoint and the CASRAM disclosure newly bring into the argument.}

\begin{center}\rule{0.5\linewidth}{0.5pt}\end{center}

\emph{Author scope note (not for publication): This draft confines its
claim to the complementary/orthogonal reading. Thagard's account enters
as convergence witness and as the source of the recursion argument in
\S{}5.2 and \S{}9; it is not implemented, and no coherence-level replacement
of his theory is attempted (\S{}14, reservation iv). The Rescue-Rate
finding of \S{}7 is the one result in this draft stated as a general
constraint rather than as a property of our substrate, and \S{}14 gives
the refutation that would retire it. The companion
engineering paper {[}5{]} supplies the diffusion/QOD proofs, the CR
machinery, and the resolution-family record that \S{}12 draws on; the
companion Perspective {[}6{]} supplies the DG--CA3 reading condensed in
\S{}10. A separate paper addressed
to a cognitive-science audience would be the place to attempt the
coherence comparison this one declines. The references to CASRAM added
in this revision are disclosure of an existing product and a forthcoming
collaboration, not a validation claim; they do not upgrade any
VaCoAl-level (architectural) claim to PyVaCoAl-level (demonstrated)
status, and reviewers should hold that line regardless of how the
disclosure reads rhetorically.}

\end{document}